\documentclass[a4paper, 10pt, conference]{IEEEtran}
\IEEEoverridecommandlockouts
\usepackage{cite}
\usepackage{amsmath,amssymb,amsfonts}
\usepackage{algorithm}
\usepackage{algorithmic}
\usepackage[dvips]{graphicx}
\usepackage{textcomp}
\usepackage{xcolor}
\def\BibTeX{{\rm B\kern-.05em{\sc i\kern-.025em b}\kern-.08em
    T\kern-.1667em\lower.7ex\hbox{E}\kern-.125emX}}

\usepackage{subfigure}
\usepackage{multirow}
\usepackage{diagbox}
\usepackage{url}

\newcommand{\bvec}[1]{\mbox{\boldmath $#1$}}

\begin{document}


\title{An Embedded System for Image-based Crack Detection by using Fine-Tuning model of Adaptive Structural Learning of Deep Belief Network
\thanks{\copyright 2020 IEEE. Personal use of this material is permitted. Permission from IEEE must be obtained for all other uses, in any current or future media, including reprinting/republishing this material for advertising or promotional purposes, creating new collective works, for resale or redistribution to servers or lists, or reuse of any copyrighted component of this work in other works.}
}
\author{
\IEEEauthorblockN{Shin Kamada}
\IEEEauthorblockA{Advanced Artificial Intelligence Project Research Center,\\
Research Organization of Regional Oriented Studies,\\
Prefectural University of Hiroshima\\
1-1-71, Ujina-Higashi, Minami-ku, \\
Hiroshima 734-8558, Japan\\
E-mail: skamada@pu-hiroshima.ac.jp}
\and
\IEEEauthorblockN{Takumi Ichimura}
\IEEEauthorblockA{Advanced Artificial Intelligence Project Research Center,\\
Research Organization of Regional Oriented Studies,\\
Faculty of Management and Information System,\\
Prefectural University of Hiroshima\\
1-1-71, Ujina-Higashi, Minami-ku, \\
Hiroshima 734-8558, Japan\\
E-mail: ichimura@pu-hiroshima.ac.jp}
}

\maketitle

\begin{abstract}
  Deep learning has been a successful model which can effectively represent several features of input space and remarkably improve image recognition performance on the deep architectures. In our research, an adaptive structural learning method of Restricted Boltzmann Machine (Adaptive RBM) and Deep Belief Network (Adaptive DBN) have been developed as a deep learning model. The models have a self-organize function which can discover an optimal number of hidden neurons for given input data in a RBM by neuron generation-annihilation algorithm, and can obtain an appropriate number of RBM as hidden layers in the trained DBN. The proposed method was applied to a concrete image benchmark data set SDNET 2018 for crack detection. The dataset contains about 56,000 crack images for three types of concrete structures: bridge decks, walls, and paved roads. The fine-tuning method of the Adaptive DBN can show 99.7\%, 99.7\%, and 99.4\% classification accuracy for test dataset of three types of structures. In this paper, our developed Adaptive DBN was embedded to a tiny PC with GPU for real-time inference on a drone. For fast inference, the fine tuning algorithm also removed some inactivated hidden neurons to make a small model and then the model was able to improve not only classification accuracy but also inference speed simultaneously. The inference speed and running time of portable battery charger were evaluated on three kinds of Nvidia embedded systems; Jetson Nano, AGX Xavier, and Xavier NX.
\end{abstract}

\begin{IEEEkeywords}
Adaptive Structural Learning, Deep Belief Network, Deep Learning, Crack Detection, SDNET 2018, Embedded System
\end{IEEEkeywords}

\section{Introduction}
In recent years, Artificial Intelligence (AI) related technology has shown remarkable development centering on deep learning. With major achievements in image recognition and speech recognition, AI continues to disrupt society because recent deep learning has exploded with interesting and promising results. In addition to developing deep learning models with high originality, large amounts of data are collected to find regularity and relevance, and to make judgment and prediction by using the developed models in the real world problems.

In our previous research, the adaptive structural learning method of Deep Belief Network (Adaptive DBN) \cite{Kamada18_Springer} has been developed, which has an outstanding function to find the optimal network structure for given input. Since DBN is a stacking model of pre-trained Restricted Boltzmann Machines (RBM) \cite{Hinton06, Hinton12}, the proposed method can firstly determine an optimal number of neurons for one RBM by hidden neuron generation and annihilation algorithm, and then optimal layers of RBM are also generated according to the total error during the training of deep learning model. Adaptive DBN method shows the highest classification capability for image recognition of some benchmark data sets such as MNIST \cite{LeCun98a}, CIFAR-10, and CIFAR-100 \cite{CIFAR10}. The paper \cite{Kamada18_Springer} reported the model can reach higher classification accuracy for test cases than existing Convolutional Neural Network (CNN) such as AlexNet \cite{AlexNet}, GoogLeNet \cite{GoogLeNet}, VGG16 \cite{VGG16}, and ResNet \cite{ResNet}.

Recently, more sophisticated applications with AI has been also required for civil engineering in construction management, design optimization, risk control, and quality control. For example, the immediate maintenance and inspection operation for a road structure, bridge, debris barrier, and so on, was required after extraordinary weather such as torrential rain and typhoon. The inspection for concrete bridge is not easy operation, because the bridge inspection car should have special equipment such as ladder to watch the surface at a high altitude. Thus it is expected to use a drone or UAV system for the inspection of concrete structure to realize safety and efficiency system. Moreover, the experts for crack detection are decreasing, since building up the expert knowledge takes a long term education. The deep learning technology for crack detection in the field of civil engineering is required to solve the problems.

SDNET 2018 \cite{SDNET2018} is an annotated concrete image dataset for crack detection algorithms. It consists of over 56,000 cracked and non-cracked images for three types of structure; concrete bridge decks, walls, and pavements. They are used for training of crack detection algorithms such as deep learning \cite{CNN_CONCRETE}. The dataset includes images with many kinds of obstructions such as shadows, surface roughness, scaling, edges, holes, and background debris. SDNET 2018 will be one of benchmark dataset in the field of structural health monitoring and the concrete crack detection method have been developed by using some deep learning models \cite{Cha2017}. In \cite{SDNET2018}, the classification accuracy was reported 91.9\%, 89.3\%, and 95.5\% for three types of structures as the results of transfer learning by using the AlexNet \cite{AlexNet}.

The objective of our research is to develop an image based deep learning system which can realize fast concrete crack detection using a drone. For a deep learning model, our Adaptive DBN trained SDNET 2018 and it showed 96.5\%, 96.8\%, and 96.5\% classification accuracy and the values are higher than the result reported in \cite{SDNET2018}. After training of Adaptive DBN, the fine-tuning method described in \cite{Kamada17_IJCIStudies} can show 99.7\%, 99.7\%, and 99.4\%, respectively. 

In this paper, our developed Adaptive DBN was embedded to a tiny PC with GPU for real-time inference on a drone. The NVIDIA Jetson series are known to be a tiny embedded system which enables fast inference of deep learning. For faster inference, the fine-tuning algorithm also found some inactivated hidden neurons and removed them to make the model size small. Generally, when reducing the trained model size such as a distillation learning, classification accuracy and inference speed become a relation of trade-off, but the fine tuning was able to improve not only classification accuracy but also inference speed simultaneously. We evaluated our developed model on three kinds of Jetson; Jetson Nano, AGX Xavier, and Xavier NX.

The remainder of this paper is organized as follows. In section \ref{sec:adaptive_dbn}, the basic idea of the adaptive structural learning of DBN is briefly explained. In section \ref{sec:SDNET2018}, the effectiveness of our proposed method is verified on SDNET 2018. Section \ref{sec:embedded_system} give a description about our developed embedded system and the inference speed and running time of portable battery charger are reported. In section \ref{sec:conclusion}, some discussions are given to conclude this paper.

\section{Adaptive Structural Learning Method of Deep Belief Network}
\label{sec:adaptive_dbn}
In this section, the brief description about our proposed Adaptive DBN is explained to understand the basic behavior of self-organized structure.

\subsection{RBM and DBN}
\label{sec:RBMDBN}
\begin{figure}[bt]
\centering
\includegraphics[scale=0.8]{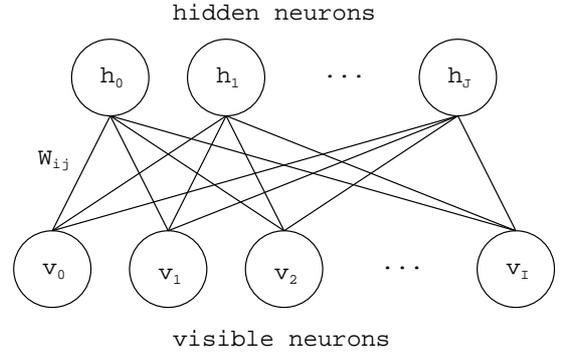}
\caption{Network structure of RBM}
\label{fig:rbm}
\end{figure}

A RBM \cite{Hinton12} is a kind of stochastic model for unsupervised learning with two kinds of binary layers as shown in Fig.~\ref{fig:rbm}: a visible layer $\bvec{v} \in \{0, 1 \}^{I}$ for input patterns and a hidden layer $\bvec{h} \in \{0, 1 \}^{J}$. The parameters $\bvec{b} \in \mathbb{R}^{I}$ and $\bvec{c} \in \mathbb{R}^{J}$ can work to make an adjustment of behaviors for a visible neuron and a hidden neuron, respectively. The weight $W_{ij}$ is the connection between a visible neuron $v_{i}$ and a hidden neuron $h_j$. A RBM minimizes the energy function $E(\bvec{v}, \bvec{h})$ during training in the following equations.
\begin{equation}
E(\bvec{v}, \bvec{h}) = - \sum_{i} b_i v_i - \sum_j c_j h_j - \sum_{i} \sum_{j} v_i W_{ij} h_j ,
\label{eq:energy}
\end{equation}
\begin{equation}
p(\bvec{v}, \bvec{h})=\frac{1}{Z} \exp(-E(\bvec{v}, \bvec{h})),
\label{eq:prob}
\end{equation}
\begin{equation}
Z = \sum_{\bvec{v}} \sum_{\bvec{h}} \exp(-E(\bvec{v}, \bvec{h})),
\label{eq:z}
\end{equation}
Eq.~(\ref{eq:prob}) is a probability of $\exp(-E(\bvec{v}, \bvec{h}))$. $Z$ is calculated by summing energy for all possible pairs of visible and hidden vectors in Eq.~(\ref{eq:z}). The parameters $\bvec{\theta}=\{\bvec{b}, \bvec{c}, \bvec{W} \}$ is optimized by partial derivative of $p(\bvec{v})$ for given input data $\bvec{v}$ . A standard estimation for the parameters is maximum likelihood in a statistical model $p(\bvec{v})$.

DBN \cite{Hinton06} is a hierarchical stacking model using several pre-trained RBMs. The activation values of hidden neurons at $l-1$-th RBM are taken to next level of input at $l$-th RBM as Eq.~(\ref{eq:prob_dbn}).

\begin{equation}
\label{eq:prob_dbn}
p(h_j^{l} = 1 | \bvec{h}^{l-1})= sigmoid(c^{l}_j + \sum_{i}W^{l}_{ij} h^{l-1}_{i}),
\end{equation}
where $c^{l}_j$ and $W^{l}_{ij}$ indicate the parameters for the $j$-th hidden neuron and the weight at the $l$-th RBM, respectively. $\bvec{h}^{0} = \bvec{v}$ means the given input data. In order to make the DBN a supervised learning for a classification task, the last layer appends to the final output layer to calculate the output probability $y_k$ for a category $k$. The calculation is implemented by Softmax in Eq.(\ref{eq:softmax}). 
\begin{equation}
\label{eq:softmax}
y_k = \frac{\exp(z_{k})}{\sum^{M}_{j} \exp(z_j)},
\end{equation}
where $z_{j}$ is an output pattern of the $j$-th hidden neuron at the output layer. The number of output neurons is $M$. The difference between the output $y_k$ and the teacher signal for the category $k$ is minimized. 

\subsection{Neuron Generation and Annihilation Algorithm of RBM}
\label{subsec:adaptive_rbm}
Recently, deep learning models treat large amount of big data to realize higher classification performance. On the other hand, the size of its network structure or the number of its parameters that a network designer determines must become larger. In our research, to solve such a problem, the adaptive structural learning method of RBM, Adaptive RBM \cite{Kamada18_Springer}, was developed to find the optimal number of hidden neurons for given input space during its training situation. The key idea of Adaptive RBM is the observation of Walking Distance (WD), which means a criterion how the model's parameters are changed on the iterative learning \cite{Kamada16_SMC}. If there are not enough neurons to classify the input images in the RBM, then WD may fluctuate even after a long training. We selected to monitor two parameters ($\bvec{c}$ and $\bvec{W}$) of $\bvec{\theta}=\{\bvec{b}, \bvec{c}, \bvec{W} \}$ which have influenced on the learning convergence of RBM except the parameter, $\bvec{b}$, related to input data. Due to a cross check system, please see the paper \cite{Kamada18_Springer} for detailed algorithm of the method.

\begin{figure}[]
\begin{center}
\subfigure[Neuron generation]{\includegraphics[scale=0.5]{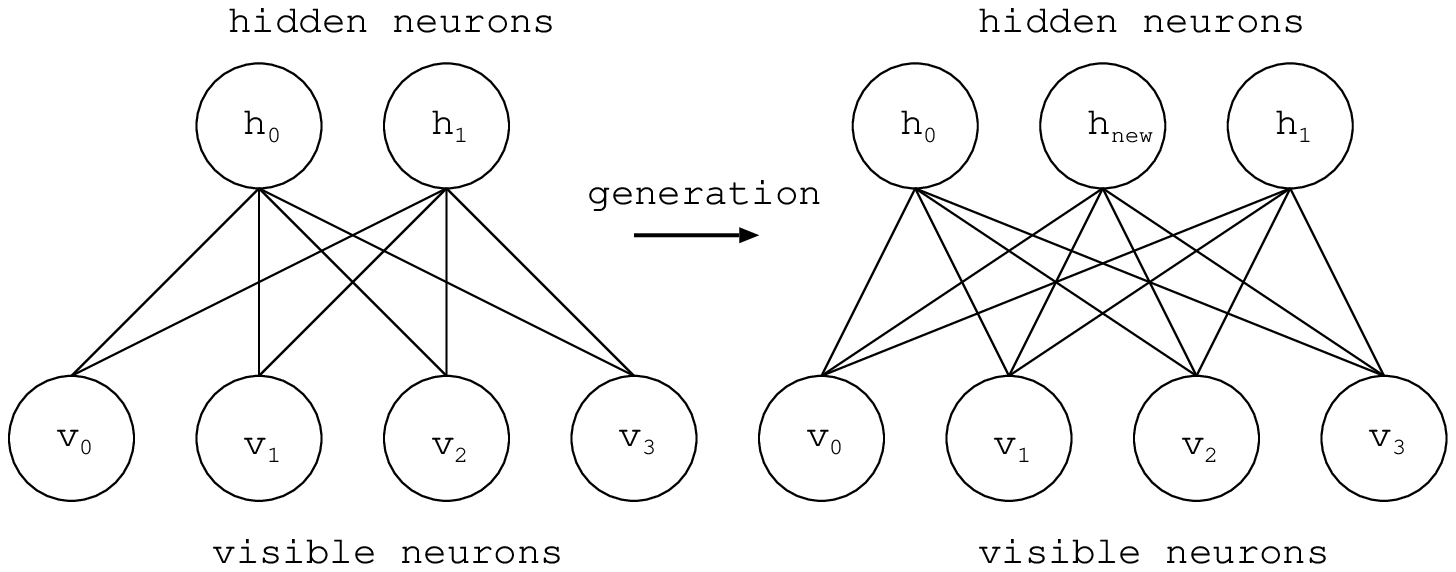}\label{fig:neuron_generation}}
\subfigure[Neuron annihilation]{\includegraphics[scale=0.5]{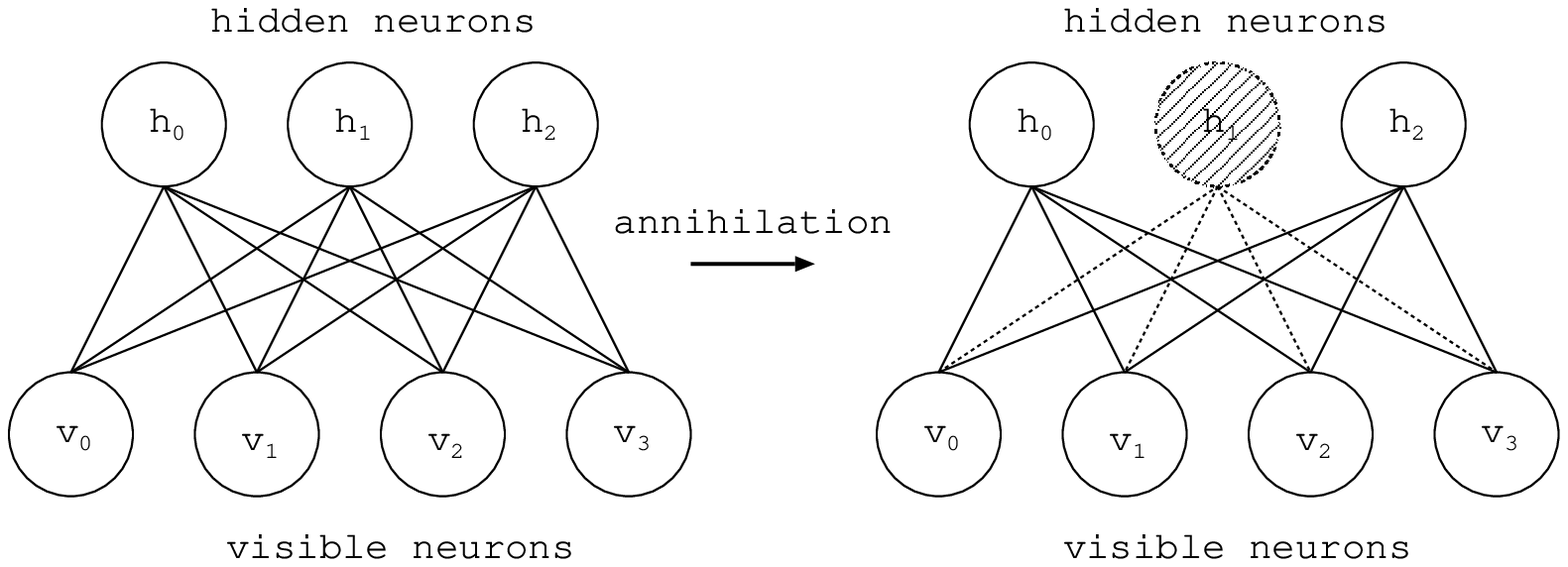}\label{fig:neuron_annihilation}}
\vspace{-3mm}
\caption{Adaptive RBM}
\label{fig:adaptive_rbm}
\end{center}
\end{figure}

However, a situation is supposed that some redundant neurons will be generated due to the neuron generation process. The neuron annihilation algorithm is operated to kill the corresponding neuron after neuron generation process. Fig.~\ref{fig:neuron_annihilation} shows that the corresponding neuron is annihilated \cite{Kamada16_ICONIP}.

\begin{figure*}[]
\centering
\includegraphics[scale=0.8]{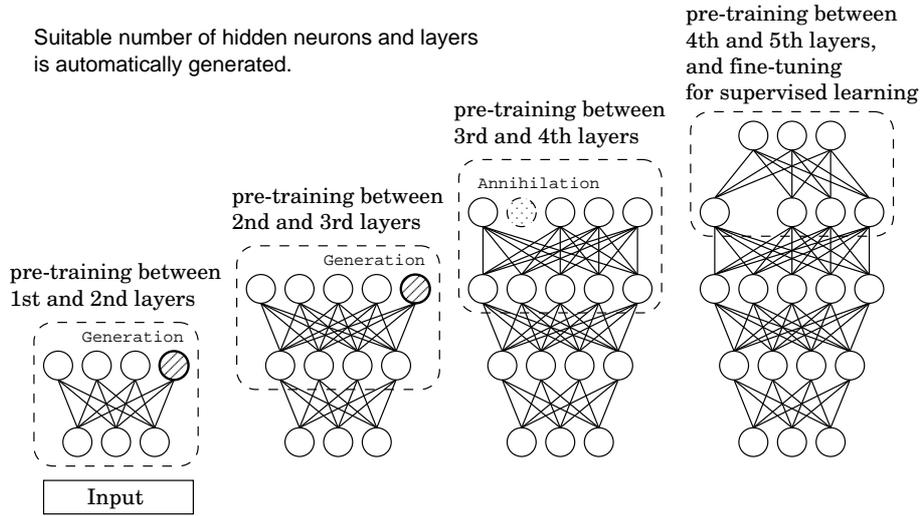}
\vspace{-3mm}
\caption{An Overview of Adaptive DBN}
\label{fig:adaptive_dbn}
\end{figure*}

\subsection{Layer Generation Algorithm of DBN}
\label{subsec:adaptive_dbn}
Fig.~\ref{fig:adaptive_dbn} shows the overview of layer generation in Adaptive DBN. Our DBN model is a generative neural network model with hierarchical layered of the two or more pre-trained RBMs. The input of $l+1$-th RBM is given by the output of $l$-th RBM. DBN with many layered RBMs will have higher classification capability of data representation. Such hierarchical model can represent the various features from the lower-level features to concrete image in the direction of input layer to output layer. In order to realize the sophisticated learning, the optimal number of RBMs depends on the target data space \cite{Kamada16_TENCON}.

Our developed Adaptive DBN can accomplish to make an optimal network structure to automatically add the RBM one by one based on the idea of WD described in section \ref{subsec:adaptive_rbm}. If the energy values and WD don't converge, each RBM lacks presentation capability to classify input patterns. In order to solve such a situation, the refined network structure is self-organized by a new generated RBM to obtain higher classification power \cite{Kamada18_Springer}. Eqs.~(\ref{eq:layer_generation1})-(\ref{eq:layer_generation2}) are the condition for layer generation is defined by using the WD of RBM and the energy function as follows.

\begin{equation}
\sum_{l=1}^{k} WD^{l} > \theta_{L1},
\label{eq:layer_generation1}
\vspace{-3mm}
\end{equation}
\begin{equation}
\sum_{l=1}^{k} E^{l} > \theta_{L2}
\label{eq:layer_generation2}
\end{equation}

\section{SDNET 2018}
\label{sec:SDNET2018}
\subsection{Data Description}
\label{sec:Data_Description}
\begin{figure}[btp]
  \centering
  \subfigure[Bridge deck, w/o cracks]{\includegraphics[scale=0.3]{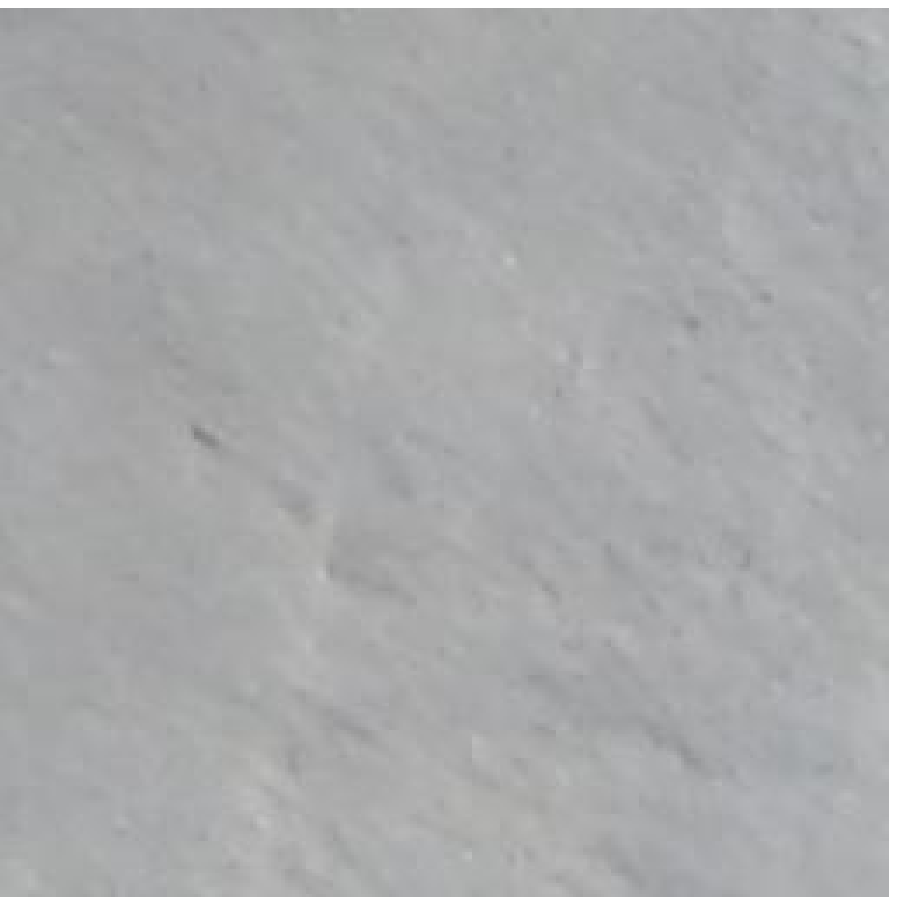}\label{fig:UD}}
  \subfigure[Bridge deck, with cracks]{\includegraphics[scale=0.3]{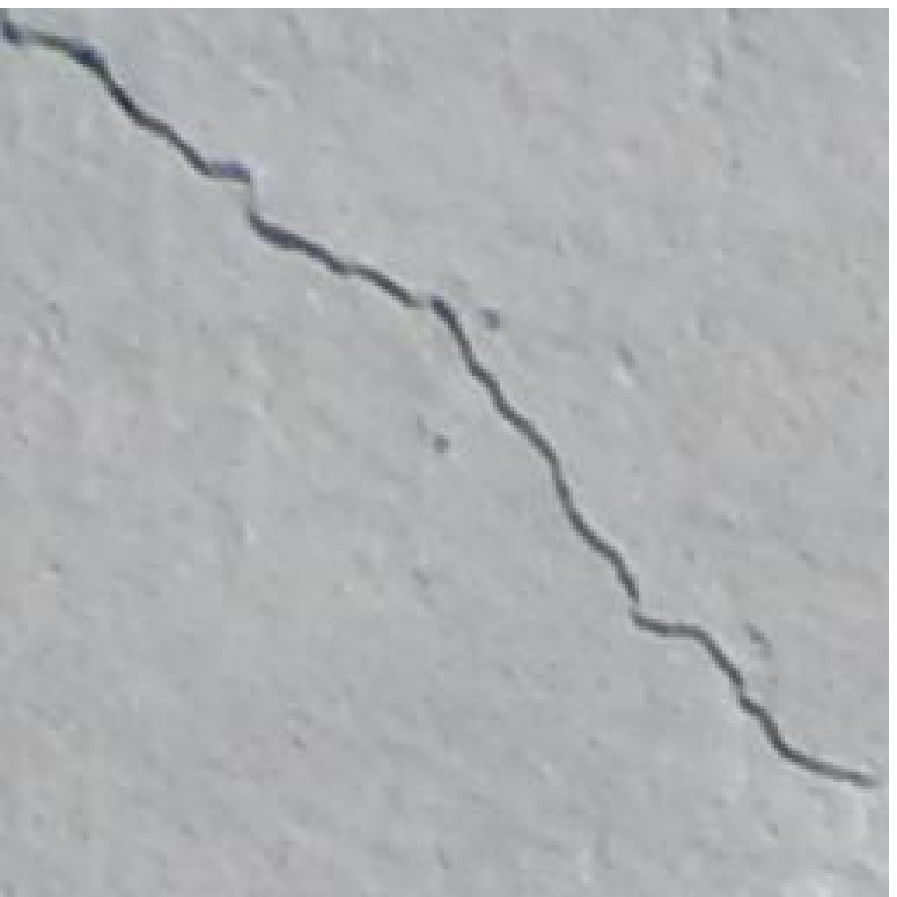}\label{fig:CD}}
  \subfigure[Wall, w/o cracks]{\includegraphics[scale=0.3]{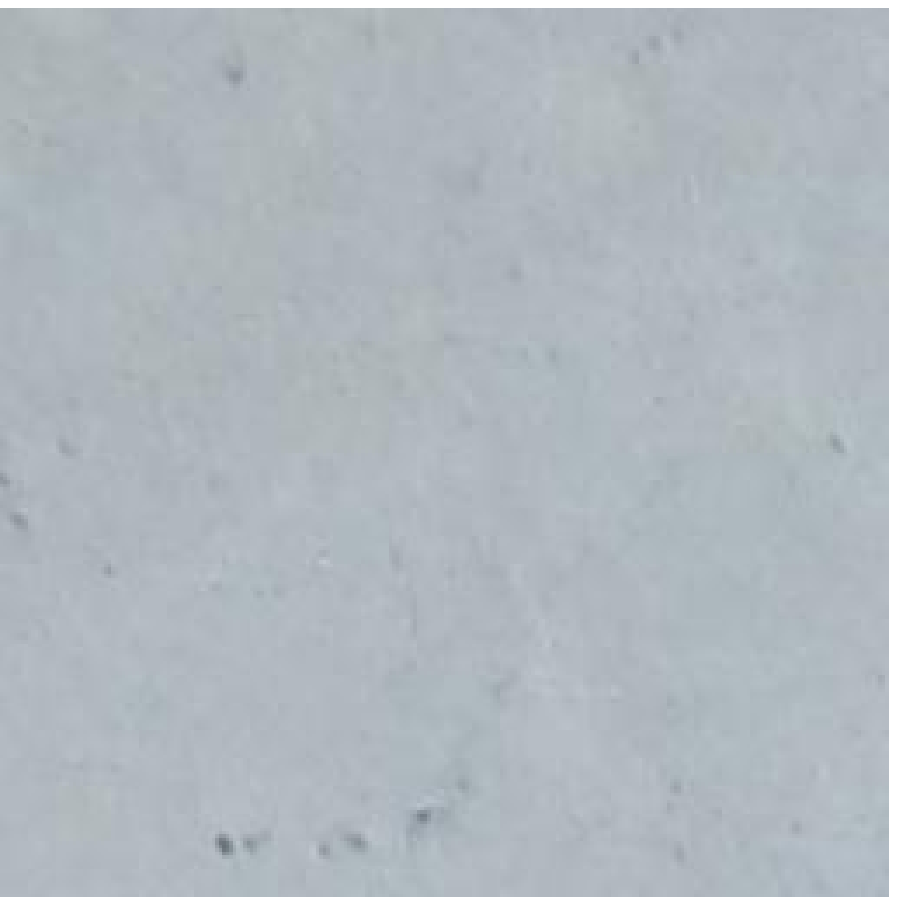}\label{fig:UW}}
  \subfigure[Wall, with cracks]{\includegraphics[scale=0.3]{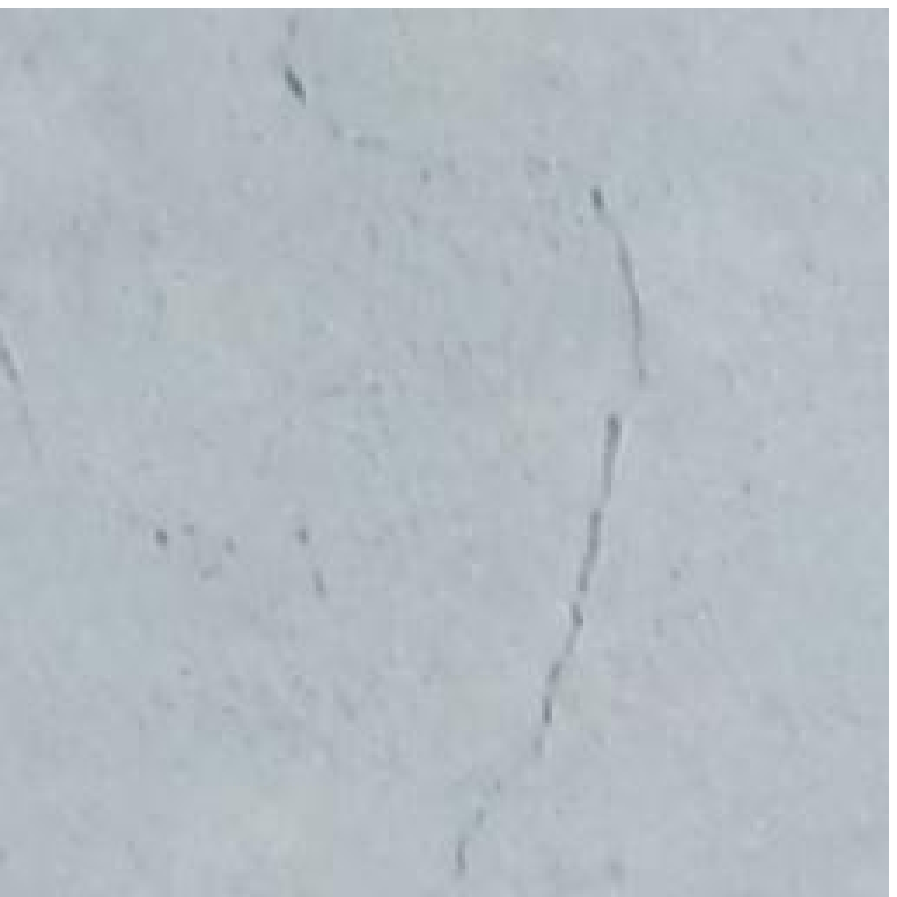}\label{fig:CW}}
  \subfigure[Pavement, w/o cracks]{\includegraphics[scale=0.3]{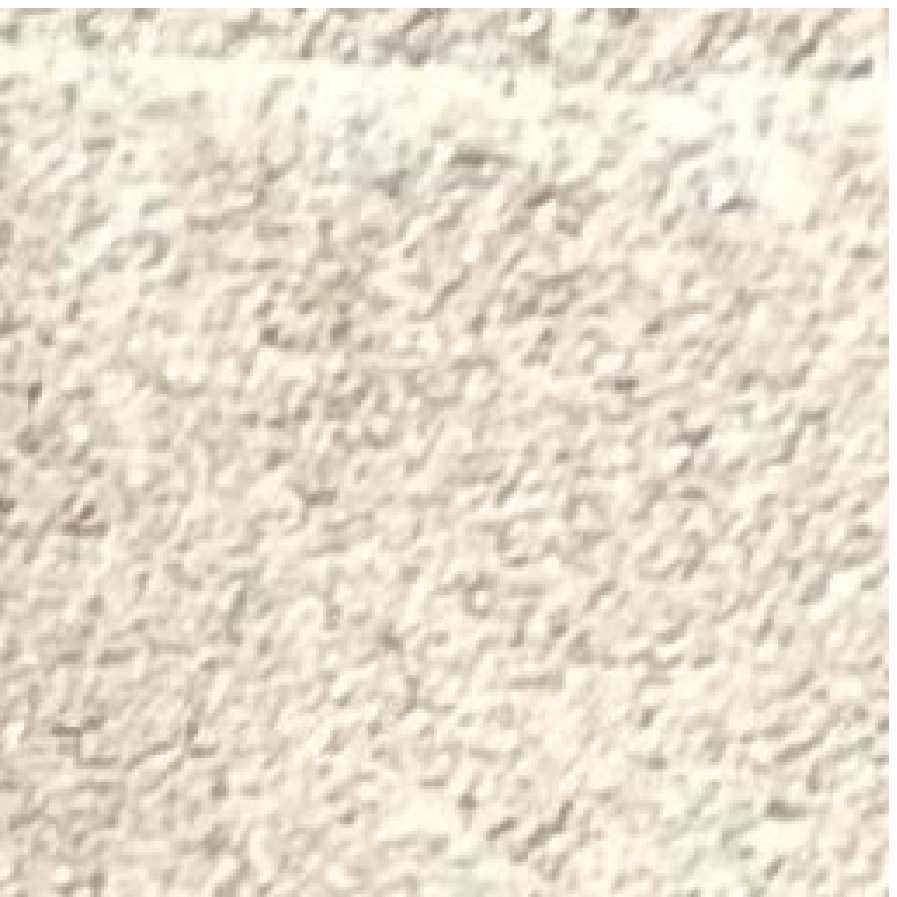}\label{fig:UP}}
  \subfigure[Pavement, with cracks]{\includegraphics[scale=0.3]{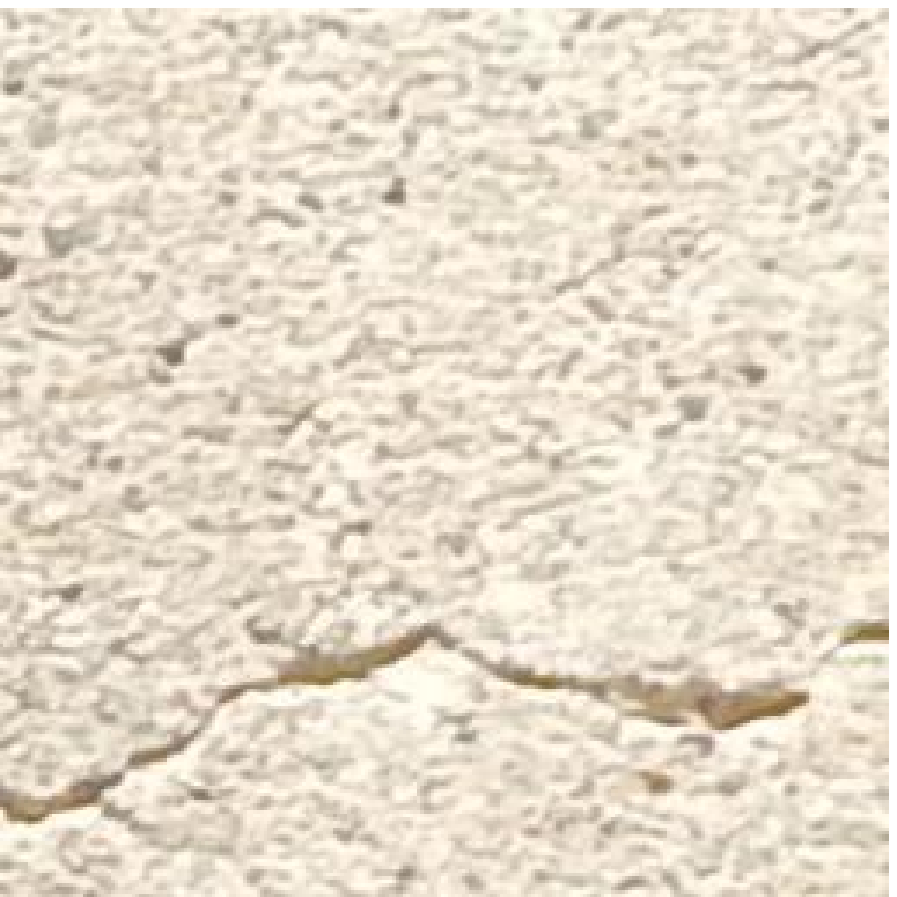}\label{fig:CP}}
  \caption{Sample of SDNET 2018}
  \label{fig:data_sdnet_sample}
\end{figure}

SDNET 2018 \cite{SDNET2018} is an annotated concrete image dataset collected from Utah State University. The dataset consists of over 56,000 cracked and non-cracked images for three types of structure; bridge decks, walls, and pavements, as a classification problem. The images are used to develop crack detection algorithm such as deep learning \cite{CNN_CONCRETE}. The dataset includes images with many kinds of obstructions such as shadows, surface roughness, scaling, edges, holes, and background debris. Table \ref{tab:data_sdnet_category} shows the number of cracked, non-cracked, and total sub-images of each type in SDNET 2018. According to the report in \cite{SDNET2018}, the dataset includes cracks at the variety size from 0.06 mm to 25 mm. Fig.~\ref{fig:data_sdnet_sample} shows the sample images of crack conditions. Each image is 256 $\times$ 256 size in colored pixel and is annotated as `with cracked' or `without cracked' for the supervised training.

\begin{table}[btp]
\caption{The category of SDNET 2018}
\vspace{-3mm}
\label{tab:data_sdnet_category}
\begin{center}
\scalebox{0.9}[0.9]{
\begin{tabular}{l|r|r}
\hline 
\multicolumn{1}{c|}{Category} & Training dataset & Test dataset  \\ 
\hline
Bridge deck w/o cracks   &  10,424  & 1,834 \\ 
Bridge deck with cracks  &     1,171  &    191 \\ \hline
Wall w/o cracks  &  12,853 & 1,434 \\ 
Wall with cracks  &    3,471 &     380 \\ \hline
Pavement w/o cracks  &  19,531  & 2,195 \\ 
Pavement with cracks  &     2,369  &    239 \\ \hline
\multicolumn{1}{c|}{total}  &     49,819  &    6,273 \\ \hline
\end{tabular}
} 
\end{center}
\end{table}

\subsection{Classification Results}
\label{sec:classification_results}
Table~\ref{tab:classification_ratio1} compares classification accuracy for SDNET 2018 test data with a conventional CNN and the Adaptive DBN. The CNN is a transfer learning based on AlexNet \cite{SDNET2018}. Our method can classify the images to cracked and non-cracked labels with accuracy of more than 95\% for three types of structures, which is higher than the existing CNN. Fine Tuning described in \cite{Kamada17_IJCIStudies} is a method to improve the classification accuracy of Adaptive DBN after learning. {\bf Algorithm 1} and {\bf Algorithm 2} show the algorithms of the fine tuning. In the model, according to frequency of input/output patterns at each layer in the network after learning, the network weights are modified so that cases classified incorrectly are classified correctly. Fine Tuning method improved the classification accuracy for test data to more than 99.4\%. 

Table~\ref{tab:classification_ratio2} shows the classification accuracy by Adaptive DBN as shown in Table~\ref{tab:classification_ratio1} for the training data and the test data for three types of structures without cracks and with cracks. The value in the parenthesis in the cell of the test data indicates ``the number of incorrect data / the total number of data''. The Adaptive DBN showed 100\% classification accuracy for training data. The test data showed classification accuracy of 95.3\% or more for all categories. The numerical values for the three types of structures were almost the same, and there was no significant difference. The numerical values did not drop significantly both with and without cracks. For about 4\% mis-classified data of test data, the detailed investigation by the experts who can judge the concrete crack images was reported \cite{Kamada20_ICIPRoB}.

\begin{table}[btp]
  \caption{Classification ratio of \cite{SDNET2018} and Adaptive DBN}
\vspace{-3mm}
\label{tab:classification_ratio1}
\begin{center}
\scalebox{0.9}[0.9]{
\begin{tabular}{l|r|r|r}
\hline
\multicolumn{1}{c|}{Category} &  \multicolumn{1}{c|}{CNN \cite{SDNET2018}} & \multicolumn{1}{c|}{Adaptive DBN} & \multicolumn{1}{c}{Fine Tuning \cite{Kamada17_IJCIStudies}} \\ \hline
Bridge deck &  91.9\%   & 96.5\% & 99.7\% \\ 
Wall              & 89.3\%  & 96.8\% & 99.7\% \\ 
Pavement   &  95.5\%  & 96.5\% & 99.4\%\\
\hline
\end{tabular}
} 
\end{center}
\vspace{-5mm}
\end{table}

\begin{figure}[]
\begin{center}
\includegraphics[width=85mm]{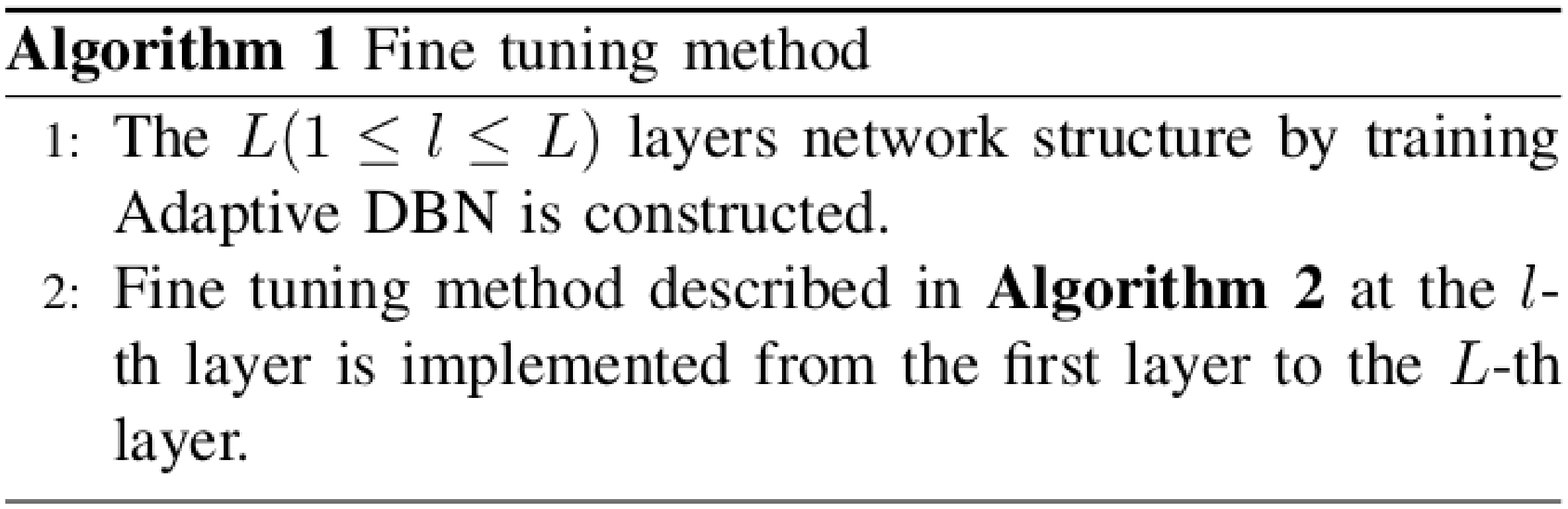}
\end{center}
\end{figure}

\begin{figure}[]
\begin{center}
\includegraphics[width=85mm]{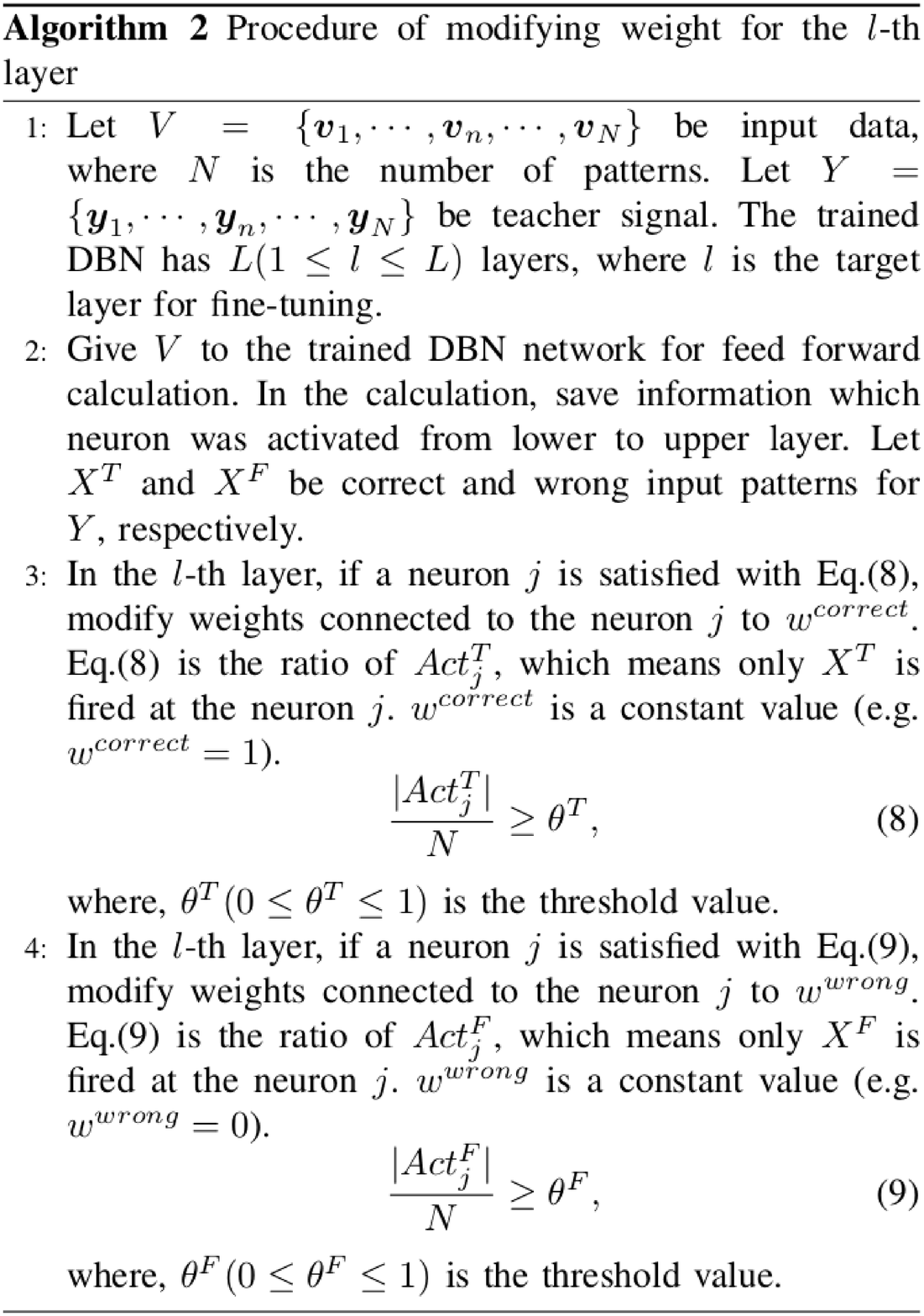}
\end{center}
\end{figure}

\begin{table}[btp]
  \caption{Classification ratio by Adaptive DBN}
\vspace{-3mm}
\label{tab:classification_ratio2}
\begin{center}
\begin{tabular}{l|r|rr}
\hline 
\multicolumn{1}{c|}{Category} & Train & \multicolumn{2}{c}{Test}  \\ 
\hline
Bridge deck w/o cracks    &  99.3\%  & 96.5\% & (64/1834)\\ 
Bridge deck with cracks   &  99.2\% & 96.3\% & (7/191) \\ \hline
Wall w/o cracks           & 98.8\%  & 97.2\% & (40/1434) \\ 
Wall with cracks          &  98.6\%  & 95.3\% & (18/380) \\ \hline
Pavement w/o cracks       &  98.2\%  & 96.6\% &  (75/2195) \\ 
Pavement with cracks      &  98.3\%  & 95.8\% &  (10/239) \\ \hline
\end{tabular}
\end{center}
\vspace{-5mm}
\end{table}

\begin{table*}[btp]
\caption{Specification of Embedded systems}
\vspace{-3mm}
\label{tab:embedded_system}
\begin{center}
\begin{tabular}{l|l|l|l}
\hline 
& Jetson Nano & Jetson AGX Xavier & Jetson Xavier NX \\ \hline
CPU & Quad-core ARM A57 & 8-core Carmel ARM v8.2 & 6-core Carmel ARM v8.2   \\ \hline
RAM & 4GB & 16GB & 8GB \\\hline
GPU & 128 Core Maxwell & 512 Core Volta & 384 Core Volta \\ \hline
Weight & 140g & 1,548g & 771g \\ \hline
Camera &  \multicolumn{3}{l}{30fps, frame size: $256 \times 256$ pixel} \\\hline
Sensor & \multicolumn{3}{l}{3 axis acceleration, gyro, magnetic field, air pressure, temperature, and GPS} \\ \hline
Portable battery & \multicolumn{3}{l}{17400mAh, 62.64Wh (12700-BTL033BK)} \\ \hline
\end{tabular}
\end{center}
\end{table*}

\begin{table*}[btp]
\caption{Inference speed of Embedded systems}
\vspace{-3mm}
\label{tab:embedded_system_result}
\begin{center}
\begin{tabular}{l|l|l|l}
\hline 
& Jetson Nano & Jetson AGX Xavier & Jetson Xavier NX  \\ \hline
Inference Speed (before fine tuning) & 88.51 ms & 28.72 ms & 37.97 ms   \\ \hline
Inference Speed (after fine tuning)  & 83.27 ms & 25.93 ms & 35.20 ms   \\ \hline
Running time of portable battery & 13 hours & 5 hours & 12 hours  \\ \hline
\end{tabular}
\end{center}
\end{table*}

\begin{table*}[tb]
\caption{Acquired network structure}
\label{tab:network_sotructure}
\begin{center}
\begin{tabular}{l|l}
\hline
                   & No. of neurons at each layer \\ \hline
Before finetuning  & $629 -> 402 -> 350 -> 301 -> 152 -> 105$ \\ 
After finetuning   & $629 -> 402 -> 349 -> 259 -> 101 -> 89$ \\ 
\hline
\end{tabular}
\end{center}
\end{table*}

\section{Embedded system}
\label{sec:embedded_system}
In this section, an embedded system with the trained network as described in section \ref{sec:classification_results}, was explained for real-time inference. In this paper, three kinds of Nvidia embedded systems, Jetson Nano, Jetson AGX Xavier, and Jetson Xavier NX, were used for comparison as shown in Table~\ref{tab:embedded_system}. It is supposed that the developed embedded system will be attached with a drone or UAV system for real-time detection of concrete structures. To realize fast detection, both of inference speed and running time of portable battery charger were compared with three kinds of embedded systems. In addition, the inference speed of the trained model was also compared with and without the fine tuning, because some inactivated neurons for inference were found by the fine tuning algorithm and then these neurons were removed for fast inference. To measure these values on same condition for three different systems, a same USB camera device was used and its frame size and fps are $256 \times 256$ and 30fps, respectively.

Table~\ref{tab:embedded_system_result} shows the inference speed and running time of portable battery charger for the three embedded systems. The inference speed means a value of averaged inference mill second per one image. The running time of portable battery means a time that the fully charged battery is dead. From Table~\ref{tab:embedded_system_result}, Jetson AGX Xavier had the highest inference speed, but the running time of portable battery was the lowest among the three embedded systems. On the other hand, Jetson Nano showed the lowest inference speed, but its running time of portable battery was twice higher than Jetson AGX Xavier.

Table~\ref{tab:network_sotructure} shows the change of the network structure by the fine tuning. The values in Table~\ref{tab:network_sotructure} means the number of hidden neurons at each layer to output except input and output layers. By the fine tuning, about 100 hidden neurons were removed at the higher layer (close to output) because these neurons were not used for inference in the dataset. On the other hand, the deletion of the hidden neurons was not almost occurred at the lower layer (close to input). Since DBN is a stacking model using pre-trained RBMs, the higher layer represents more concrete features of input patterns by building the lower layer's abstract features. We consider some concrete features became unnecessary in the combination process of abstract features. By reducing the trained model's size, the inference speed was slightly higher (about 5\%) than before the fine tuning. To make a further smaller model, we can use a distillation learning although the classification accuracy may slightly decrease. However, the fine tuning has an advantage to improve not only classification accuracy but also inference speed simultaneously. FPS for Nano, AGX Xavier, and Xavier NX, were 11.2, 34.8, and 27.0 before the fine tuning, 12.1, 38.5, and 28.4 after the fine tuning.

Of course, the inference speed of these embedded systems are not faster than a desktop GPU workstation (for information, FPS was 39.6 when using a desktop PC with RTX 2080 Ti, Intel(R) Core(TM) i7-7800X CPU @ 3.50GHz, 64GB RAM). However, Jetson AGX Xavier and Xavier NX had the inference speed with about 30 FPS, they can be used for real time-detection of concrete structures on a drone.

Jetson Xavier NX will be attached to a drone (we have already selected, but we cannot describe the specification due to confidentiality), because Jetson Xavier NX has higher inference performance with smaller size and weight among them. Then a demonstration experiment of the developed system will be conducted for Japanese concrete structures. Moreover, the result of real-time detection will be visualized on a tablet device, which can help a inspector to understand degree of damage and easily collect concrete data. Finally, the collected concrete images will be published as a benchmark dataset for research purpose, as well as SDNET 2018.

\section{Conclusive Discussion}
\label{sec:conclusion}
We focused on RBM and DBN, which are statistical models using the concept of likelihood, and proposed a structure-adaptive DBN that finds the optimal structure by generating / erasing neurons and hierarchization during learning. In this paper, the proposed model was applied to open data SDNET 2018 on cracks in concrete structures. As a result of the experiment, it was possible to classify the test data of the three types of structures with higher accuracy than the CNN. Moreover, an embedded system with the trained network on the jetson interface was developed for real-time inference using a drone. The fine tuning method made the trained model smaller and then the small model enabled faster inference with higher classification accuracy. In future, a demonstration experiment of the developed system with a drone will be conducted for Japanese concrete structures.

\section*{Acknowledgment}
This work was supported by JSPS KAKENHI Grant Number 19K12142, 19K24365, and the obtained from the commissioned research by National Institute of Information and Communications Technology (NICT, 21405), JAPAN.


\begin{thebibliography}{00}
\bibitem{Kamada18_Springer}
S.Kamada, T.Ichimura, A.Hara, and K.J.Mackin, \emph{Adaptive Structure Learning Method of Deep Belief Network using Neuron Generation-Annihilation and Layer Generation}, Neural Computing and Applications, pp.1--15 (2018)

\bibitem{Hinton06}
G.E.Hinton, S.Osindero and Y.Teh, \emph{A fast learning algorithm for deep belief nets}, Neural Computation, vol.18, no.7, pp.1527--1554 (2006)
  
\bibitem{Hinton12}
 G.E.Hinton, \emph{A Practical Guide to Training Restricted Boltzmann Machines}, Neural Networks, Tricks of the Trade, Lecture Notes in Computer Science (LNCS, vol.7700), pp.599--619 (2012)

\bibitem{LeCun98a}
Y.LeCun, L.Bottou, Y.Bengio, and P.Haffner, \emph{Gradient-based learning applied to document recognition}, Proc. of the IEEE, vol.86, no.11, pp.2278--2324 (1998)
  
\bibitem{CIFAR10}
A.Krizhevsky: \emph{Learning Multiple Layers of Features from Tiny Images}, Master of thesis, University of Toronto (2009)

\bibitem{AlexNet}
A.Krizhevsky, I.Sutskever, G.E.Hinton, \emph{ImageNet Classification with Deep Convolutional Neural Networks}, Proc. of Advances in Neural Information Processing Systems 25 (NIPS 2012) (2012)

\bibitem{GoogLeNet}
C.Szegedy, W. Liu, Y.Jia, P.Sermanet, S.Reed, D.Anguelov, D.Erhan, V.Vanhoucke, A.Rabinovich, \emph{Going Deeper with Convolutions}, Proc. of CVPR2015 (2015)

\bibitem{VGG16}
K.Simonyan, A.Zisserman, \emph{Very deep convolutional networks for large-scale image recognition}, Proc. of International Conference on Learning Representations (ICLR 2015) (2015)

\bibitem{ResNet}
K.He, X.Zhang, S.Ren, J.Sun, J, \emph{Deep residual learning for image recognition}, Proc. of 2016 IEEE Conference on Computer Vision and Pattern Recognition (CVPR), pp.770--778 (2016)
  
\bibitem{SDNET2018}
S.Dorafshan, R.J.Thomas, and M.Maguire, \emph{SDNET2018: An annotated image dataset for non-contact concrete crack detection using deep convolutional neural networks}, Data in Brief, vol.21, pp.1664--1668, doi.org/10.1016/j.dib.2018.11.015 (2018)
  
\bibitem{CNN_CONCRETE}
S. Dorafshan, R.J.Thomas, M.Maguire, \emph{Comparison of deep convolutional neural networks and edge detectors for image-based crack detection inconcrete}, Constr. Build. Mater. vol.186, pp.1031--1045(2018)

\bibitem{Cha2017}
YJ.Cha, W.Choi, O.Bykztrk, \emph{Deep Learning-Based Crack Damage Detection Using Convolutional Neural Networks}, Computer-Aided Civil and Infrastructure Engineering, vol.32, no.5, pp.361--378 (2017)

\bibitem{Kamada17_IJCIStudies}x
S. Kamada and T.Ichimura, \emph{Fine Tuning of Adaptive Learning of Deep Belief Network for Misclassification and its Knowledge Acquisition}, International Journal Computational Intelligence Studies, vol.6, no.4, pp.333--348 (2017)

\bibitem{Kamada16_SMC}
S.Kamada and T.Ichimura, \emph{An Adaptive Learning Method of Restricted Boltzmann Machine by Neuron Generation and Annihilation Algorithm}. Proc. of 2016 IEEE International Conference on Systems, Man, and Cybernetics (SMC2016), pp.1273--1278 (2016)  
  
\bibitem{Kamada16_ICONIP}
S.Kamada, T.Ichimura, \emph{A Structural Learning Method of Restricted Boltzmann Machine by Neuron Generation and Annihilation Algorithm}, Neural Information Processing, Proc. of the 23rd International Conference on Neural Information Processing, Springer LNCS9950), pp.372--380 (2016)

\bibitem{Kamada16_TENCON}
S.Kamada and T.Ichimura, \emph{An Adaptive Learning Method of Deep Belief Network by Layer Generation Algorithm}, Proc. of IEEE TENCON2016, pp.2971--2974 (2016)
     
\bibitem{Kamada20_ICIPRoB}
S.Kamada, T.Ichimura, and T.Iwasaki, \emph{An Adaptive Structural Learning of Deep Belief Network for Image-based Crack Detection in Concrete Structures Using SDNET2018}, Proc. of 2020 International Conference on Image Processing and Robotics (ICIPRoB 2020), pp.1--6 (2020)
  
\end{thebibliography}
\end{document}